\documentclass[letterpaper, 10 pt, journal, twoside]{IEEEtran}
\usepackage{algpseudocode}
\usepackage{algorithm}
\usepackage{graphicx}

\usepackage{rotating}
\usepackage{tikz}
\usepackage{ragged2e}
\usepackage{amssymb}
\usetikzlibrary{positioning}
\usepackage{amsmath}
\usetikzlibrary{decorations.pathreplacing}
\usepackage{tabularx}
\usepackage{url} % added by Roderich
\usepackage{graphicx}
\usepackage{subfig}   % IEEE-approved replacement for subfigure
\usepackage{cite}
\usepackage{changes}
\usepackage[
    colorlinks=true,
    linkcolor=blue,
    citecolor=blue,
    urlcolor=blue,
]{hyperref}
\usepackage{booktabs}
\usepackage{makecell}
\usepackage{multirow}
\usepackage{array}
\newcolumntype{P}[1]{>{\centering\arraybackslash}p{#1}}

\usepackage{geometry}
\IEEEoverridecommandlockouts                              % This command is only needed if 
\title{\LARGE \bf Bridging Stability and Flexibility in Deformable Object Robotics Grasping across FEM, SPH and Spring-Regularized SPH Models}

\author{
Leyi Ye$^{1}$, Stanley C. Obilikpa$^{2}$, and John Oyekan$^{3}$% <-this % stops a space
\thanks{$^{1}$Leyi Ye is with the Department of Computer Science, The University of York, UK}
\thanks{$^{2}$S.C. Obilikpa are with the School of Electrical and Electronic Engineering, The University of Sheffield, UK
        {\tt\small scobilikpa1@sheffield.ac.uk}}%
\thanks{$^{3}$J. Oyekan is with the Wolfson School of Mechanical, Electrical and Manufacturing Engineering, Loughborough University, UK}
\thanks{Code and datasets will be released to support reproducibility.}}
\begin{document}

\maketitle
\thispagestyle{empty}
\pagestyle{empty}

\begin{abstract}
Accurate simulation of deformable objects is important for robotic manipulation tasks such as food handling and packaging. The Finite Element Method (FEM) is widely used for solid mechanics and grasp simulation, but its fixed mesh connectivity makes extreme deformation, separation and rupture difficult to represent. Smoothed Particle Hydrodynamics (SPH), by contrast, is mesh-free and naturally accommodates large deformations and complex contact. However, in grasping it can become overly compliant and lose cohesion. In this work, we compare three approaches: \textbf{(i)}FEM represented by DefGraspSim; \textbf{(ii)} SPH implementations in PySPH and Taichi and \textbf{(iii)} a spring-regularized SPH variant (SPH+MSM) that overlays a sparse mass-spring network on the particle set to improve shape retention during grasping. 

Across representative objects including tomato, apple, tofu, gelatin and a latex tube, we assess grasp success, deformation behaviour, sensitivity to material parameters and computational cost under a common squeeze-hold-lift protocol. The results show that FEM is the most reliable in mid-range material regimes but struggles at extreme stiffness and near-incompressibility, while baseline SPH captures deformation and slip behaviour more naturally but often fails to maintain stable load-bearing contact. The proposed SPH+MSM model substantially improves cohesion and expands the range of stable grasps, while preserving the mesh-free advantages of SPH, at the cost of additional hyperparameter calibration. The Taichi implementation also provides a substantial speed advantage over CPU-based PySPH. 

Our results show that augmenting SPH with a sparse elastic network can turn an unstable particle-based grasp model into a practical simulator for a broader range of deformable-object grasping tasks. This positions the spring-regularized SPH model as a practical middle ground between the robustness of FEM and the flexibility of mesh-free particle methods for robotic grasp simulation of deformable 3D objects.

%
% Changed from:
%For spring-based robot fabrics, both controllers perform well, whereas for rod-based robot fabrics only the open-loop controller performs well.
\end{abstract}

\section{Introduction} \label{introduction}
Robotic manipulation of deformable objects remains significantly more challenging than handling rigid objects because the shape, contact area and internal stress state evolve continuously during interaction \cite{yuanyang2025grasping} . This is especially important in applications such as food handling, packaging, healthcare and soft-material assembly, where grasp success depends not only on the pose of the object, but also on how the material deforms under contact. As a result, an accurate simulation of deformable objects is essential to testing grasping strategies, study failure modes and reducing the cost of physical experimentation. 

Among physics based modelling methods, the Finite Element Method (FEM) and Smoothed Particle Hydrodynamics (SPH) are two of the most widely used approaches \cite{arriola2020modeling} . FEM is the standard tool for solid mechanics and industrial simulations because it provides a well established framework for modelling stress-strain behaviour in deformable  bodies. It is also widely used in engineering analysis and has been adopted in robotics for grasping and soft material simulation \cite{huang2022defgraspsim} . However, FEM relies on fixed mesh connectivity, which can cause very large deformations, separation, or ruptures that are difficult to represent without additional modelling machinery. SPH by contrast, is a mesh-free particle method \cite{price2012smoothed} that naturally accommodates large deformation and complex contact without re-meshing. These properties make SPH attractive for robotic manipulation of high compliant materials. However, in contact-rich grasping SPH can become overly compliant, losing cohesion and shape retention unless carefully stabilized \cite{jana2025remedy} \cite{xiao2023penalty}.  

This trade-off motivates the present study of this paper. We compare FEM and SPH for grasping deformable objects under a common squeeze-hold-lift protocol with emphasis on behaviors that matter in robotics including pickup success, deformation realism, failure modes, parameter sensitivity and computational cost. In addition to baseline SPH, we evaluate a spring-regularised variant, denoted as SPH+MSM, in which a sparse mass-spring network is overlaid on the particle set to improve cohesion and shape retention during grasping. The spring layer is not intended to replace the SPH formulation, but rather to act as a lightweight elastic regulariser that addresses the practical weakness of the baseline particle-based solids under pinching contact. 

We validate this approach on representative objects with different compliance and geometry, including tomato, apple, tofu, gelatin and a latex tube. FEM is evaluated using DefGraspSim \cite{huang2022defgraspsim} while SPH is implemented using PySPH \cite{pypr_pysph_github} and GPU-based Taichi \cite{taichi_overview_docs} pipeline. Rather than focus only on whether an object is lifted successfully, we analyse how each method behaves across material regimes and where each one breaks down. This provides a more practically view of the usefulness of each method for deformable simulation in robotics, where a numerically stable result is not always a physically realistic one. 
The main contributions of this paper are:
\begin{itemize}
    \item A comparative evaluation of FEM and SPH for robotic grasping of 3D deformable objects under harmonised simulation conditions. 
    \item An ablation study of spring-regularised SPH model (SPH+MSM) to examine how added elastic structure affects cohesion, shape retention, and grasp stability.
    \item An analysis of method-level trade-offs in terms of stability, deformation behaviour, failure modes, runtime, and sensitivity to material parameters, together with publicly released code and datasets to support reproducibility. 
\end{itemize}

\section{Related Work} \label{Related work}
Many approaches have been researched in literature to capture deformable object manipulation. These include Finite Element Method (FEM), Smoothed Particle Hydrodynamics (SPH), Mass--spring models (MSM), learning-based methods and Hybrid particle--spring methods. 
The \textbf{Finite Element Method (FEM)}approximates a continuous body by discretizing it into a mesh of elements connected by nodes. This allows mathematical modelling of the behaviour of physical systems under various conditions of force, thermal effect or fluid dynamics \cite{nikishkov2004introduction}. By converting governing partial differential equations into a system of algebraic equations, FEM can model complex geometries, heterogeneous materials and nonlinear behaviour with accuracy that improves as the mesh is refined \cite{nikishkov2004introduction}. FEM is therefore widely used when accurate stress--strain response, contact forces, or internal deformation fields are required. Since approximations are performed locally within each element, the resulting system of equations is sparse and dependent on only nearby nodes. This makes it possible to solve problems that involve many unknown values efficiently \cite{nikishkov2004introduction}. 

In robotics and deformable object manipulation (DOM), FEM has supported the estimation of deformation, the identification of parameters and the closed-loop control of soft objects and soft robots \cite{arriola2020modeling}\cite{yin2021modeling}. For example, in \cite{sanchez2020blind}, Sanchez \textit{et al.} estimated deformation using force sensing combined with an FEM model, without relying on vision . Other approaches use FEM to compute forces consistent with desired displacements at selected control points during dexterous manipulation \cite{ficuciello2018fem}.

A key barrier to a wider deployment is the computational cost, particularly for high-resolution meshes and contact-rich interactions, where repeated solutions can be expensive and mesh quality strongly affects precision \cite{motiwale2024neural}. This has motivated reduced-order and condensed FEM pipelines, including learned condensed models that enable fast prediction and real-time embedded control while retaining a connection to mechanics \cite{navez2025modeling}. In parallel, differentiable FEM formulations have gained traction for gradient-based parameter inference, controller tuning, and design, with recent work addressing non-smooth contact and friction to obtain more reliable derivatives \cite{menager2025differentiable,si2024difftactile,gjoka2024soft,daviet2025neurally}. Despite this progress, stable and scalable handling of contact and friction remains a central challenge for real-world robotic manipulation \cite{menager2025differentiable}\cite{si2024difftactile}.

On the other hand, \textbf{Smoothed Particle Hydrodynamics (SPH)} is a mesh-free simulation method that was originally developed for fluid dynamics and later extended to deformable solids \cite{liu2010smoothed}. Instead of dividing the simulation domain into finite elements, as in FEM, SPH represents material as a set of particles that carry physical state (e.g., mass, velocity, density and constitutive variables) and interact via smoothing kernels \cite{price2012smoothed}. This formulation naturally supports large deformations, free surfaces and certain topological changes (e.g., splitting or fragmentation), which can be challenging for mesh-based methods without remeshing or specialized treatments \cite{liu2010smoothed}.

Although SPH and learning-based approaches may both use point sets, their representations differ in meaning. In SPH, particles correspond to material points governed by discretised conservation laws, yielding a physically grounded state evolution. In contrast, many learning-based DOM methods operate on point clouds as geometric surface samples and learn mappings such as “geometry + action $\rightarrow$ new geometry” without explicitly modelling forces or conservation constraints.

Recent research has improved SPH accuracy and practicality through better kernel formulations, adaptive particle refinement, and hybrid schemes that combine SPH with mesh-based solvers, aiming to increase robustness (especially near boundaries) while reducing computational cost \cite{vacondio2021grand}. SPH has also been used in interactive simulation of deformable solids and soft tissues \cite{chang2021smoothed}\cite{obilikpa2025scalable} as well as in robotics to model environments where continuum effects are central. For example, Angelidis \textit{et al.} proposed a SPH-based framework for robotic fluid interaction and then subsequently validated it against physical experiments with swimming robots \cite{angelidis2025smoothed}.  For DOM, particle representations additionally align well with modern learned dynamics models. RoboCraft, for instance, uses particle states and graph networks to learn elasto-plastic object dynamics from RGBD observations, enabling planning and control over the learned model \cite{shi2024robocraft}. Within this spectrum, SPH offers a middle ground by providing physics-based updates via conservation laws while remaining sensitive to numerical choices (particle resolution, kernel design, and boundary handling). Nevertheless, it can become computationally expensive at high resolutions \cite{angelidis2025smoothed}.

\textbf{Mass--spring models (MSM)} represent deformable bodies as point masses connected by springs (often with damping). For two connected masses $i$ and $j$ with positions $\mathbf{x}_i,\mathbf{x}_j$, rest length $L_{ij}$, stiffness $k_{ij}$, and damping $c_{ij}$, a common spring--damper force applied to $i$ is
\begin{equation}
\mathbf{f}_{ij} =
k_{ij}\left(\|\mathbf{x}_i-\mathbf{x}_j\| - L_{ij}\right)\,\mathbf{d}_{ij}
- c_{ij}\left((\mathbf{v}_i-\mathbf{v}_j)\cdot \mathbf{d}_{ij}\right)\mathbf{d}_{ij},
\end{equation}
where $\mathbf{d}_{ij} = (\mathbf{x}_i-\mathbf{x}_j)/\|\mathbf{x}_i-\mathbf{x}_j\|$ and $\mathbf{v}$ denotes velocity. Additional constraints are often introduced to prevent unrealistic over-stretching \cite{provot1995deformation}, and stability can be improved via implicit or constraint-based integration schemes \cite{baraff2023large}. MSM are popular in interactive simulation due to their simplicity and efficiency, but their behaviour depends strongly on network connectivity and parameter tuning. Moreover, spring parameters do not always map cleanly to continuum material properties, motivating variants that better approximate isotropic elasticity or provide control over effective Poisson behaviour.

\textbf{Hybrid particle--spring methods} combine these advantages with particle discretisations. A common approach is to couple SPH with spring-network structures in fluid--structure interaction settings where the spring network provides lightweight structural behaviour while the particle method handles large deformations and complex interactions \cite{monteleone2022fluid}\cite{low2023sph}. In this work, we overlay a sparse spring network onto an SPH solid as an \emph{elastic regulariser} to improve shape retention and grasp stability. We treat this SPH+MSM formulation as an ablation of the Taichi SPH pipeline, with implementation details and quantitative effects reported in Sections~\ref{methodology} and~\ref{results}.

% \subsection{Positioning of This Work}
% Motivated by the above, this work focuses on a controlled comparison of FEM and SPH for deformable object grasping scenarios. Unlike prior studies that primarily emphasize numerical accuracy in non-robotic settings (e.g., impact/cutting) or evaluate deformable object modelling methods within a single simulator/modelling assumption, we aim to quantify practical trade-offs relevant to robotics, including simulation fidelity, stability under contact, runtime, and sensitivity to material/solver parameters, under harmonized experimental conditions. We additionally study an SPH variant augmented with a sparse mass--spring regulariser (SPH+MSM) to quantify how spring-based elastic constraints affect grasp stability, deformation realism, and runtime in contact-rich manipulation.

\section{Methodology} \label{methodology}

\subsection{Objects and material parameters}

To cover the range of compliance, strength, and near‑incompressible behavior encountered in robotic manipulation, we considered five representative objects. Their nominal properties -- density ($\rho$), Young’s modulus ($E$) and Poisson’s ratio ($\nu$) -- are summarized in Table~\ref{tbl1}.

\begin{table*}[htbp]
  \centering
  \caption{Object parameters and roles in the benchmarks}
  \label{tbl1}
  \begin{tabularx}{\linewidth}{l r r r X}
    \toprule
    Object & Density (kg/m$^{3}$) & Young's modulus (Pa) & Poisson's ratio & Role / description \\
    \midrule
    Tomato     & 1000 & $5\times10^{4}$   & 0.40 & Moderately deformable with a fragile surface; used to assess surface damage and grip‑force localization. \\
    Apple      &  850 & $1.5\times10^{6}$ & 0.30 & Comparatively stiff and resilient; exercises moderate‑deformation handling. \\
    Tofu       & 1050 & $1\times10^{4}$   & 0.40 & Extremely soft, easily deformable; representative of delicate handling requiring gentle grips. \\
    Gelatine   & 1000 & $5\times10^{4}$   & 0.50 & Highly compliant, near‑incompressible response. \\
    Latex Tube &  950 & $3\times10^{6}$   & 0.49 & Thin‑walled hollow geometry to probe bending and contact‑induced deformation. \\
    \bottomrule
  \end{tabularx}
\end{table*}

Moreover, for each parameter characterizing the deformable object, five distinct values were carefully selected to cover a wide spectrum of soft material behaviors commonly encountered in robotic grasping scenarios, ensuring that simulations would reflect realistic variations of physical behaviors observed in practical applications. Unless otherwise stated, the simulations used the default parameter set
$\rho=1000~\mathrm{kg\,m^{-3}}$, $E=5\times10^{4}~\mathrm{Pa}$, and $\nu=0.3$. 
For each sensitivity study, a single parameter was varied while the remaining parameters were maintained at these defaults. Ranges and increments were chosen to reflect the realistic variability in soft materials frequently encountered in manipulation tasks.

These choices provide a broad, yet realistic coverage of soft-material behaviour and enable a controlled comparison of the methods under parameter variation.

\subsection{Simulation Frameworks}
We evaluate FEM using DefGraspSim and SPH using PySPH and Taichi. For SPH, we consider two configurations: (i) a baseline weakly-compressible SPH formulation and (ii) a spring-regularised variant (SPH+MSM) that overlays a sparse mass--spring network on the same particle set to improve cohesion and shape retention under contact.

\subsubsection{FEM Simulation}
Finite element simulations were conducted using \textbf{DefGraspSim} \cite{huang2022defgraspsim, nvlabs_defgraspsim_github}, an open-source Python framework built on \textbf{Isaac Gym} \cite{isaacgym_docs} for physics acceleration. DefGraspSim is designed specifically for robotic grasping of three-dimensional deformable objects and provides GPU-accelerated finite-element dynamics through NVIDIA’s PhysX backend.

Each object was discretized into a tetrahedral mesh to capture localized deformations and stress propagation during gripper--object interaction. The solver employs an implicit integration scheme that provides numerical stability under stiff contact conditions and large strain gradients.

For each deformable object, ten grasping trials were executed under identical boundary and control settings. The parallel-jaw gripper was closed until initial contact was detected, held in place for stabilization, and then the object was lifted vertically by \(5~\mathrm{cm}\). A trial was considered successful if the object was lifted without slip or detachment.

The simulations were executed on Ubuntu 20.04 LTS using two NVIDIA GPUs---a GeForce RTX 3090 Ti and an NVIDIA A2000. 
% This configuration enabled real-time or near-real-time computation for moderate mesh sizes, while larger or stiffer models required extended runtimes for convergence.

\subsubsection{SPH Simulation}
\begin{figure*}
    \centering
    \includegraphics[width=0.75\linewidth]{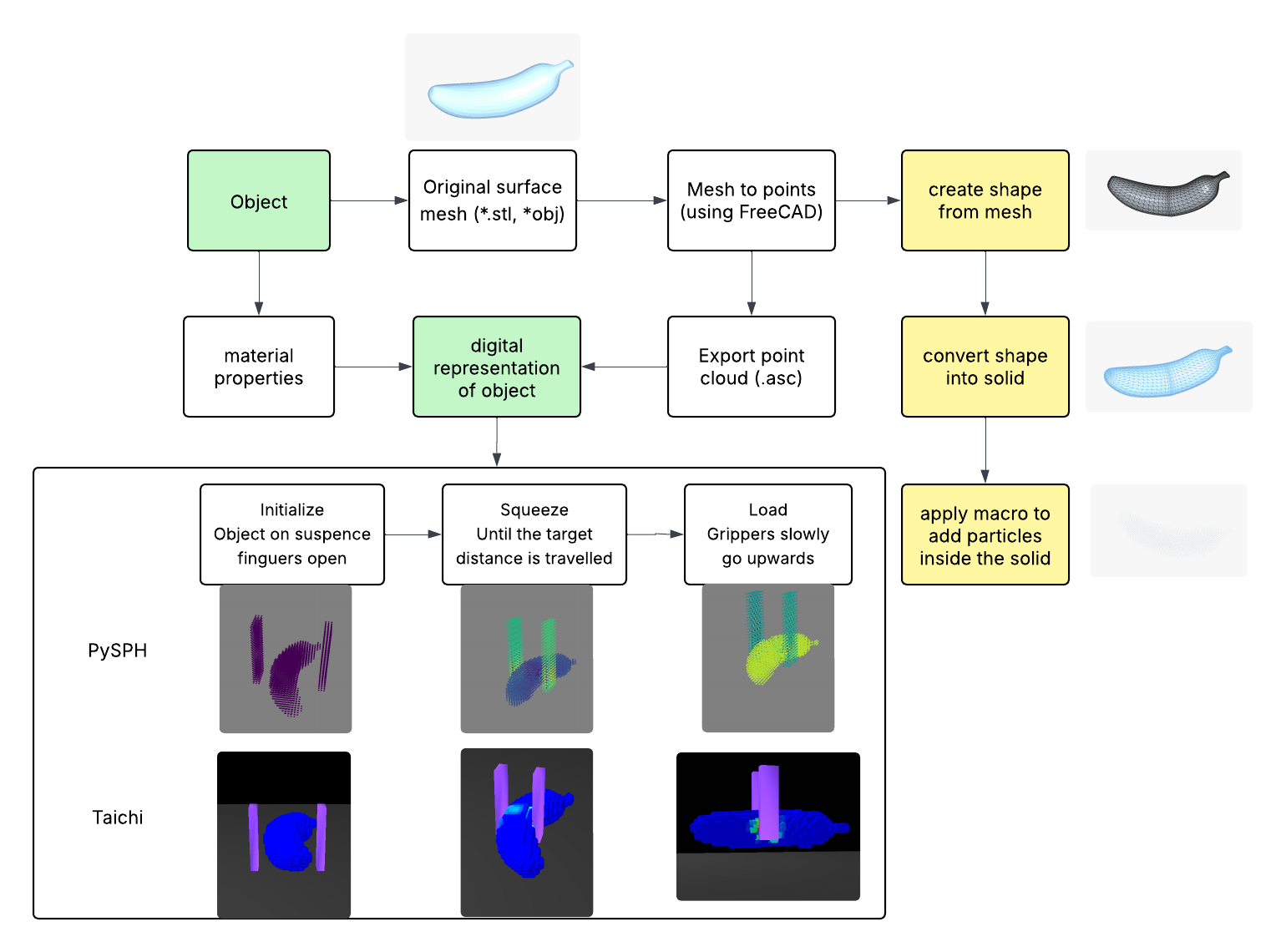}
    \caption{End-to-End processing and simulation workflow used for SPH experiments}
    \label{fig:preprocessign pipeline}
\end{figure*}

Apart from generating object point clouds directly in Python, it is also possible to initialize the SPH representation from a mesh model. Most object models available in public datasets are provided as surface meshes and must therefore be converted into a volumetric particle representation before running SPH simulations. Fig.~\ref{fig:preprocessign pipeline} summarizes this preprocessing pipeline. Starting from a watertight surface mesh (e.g., \texttt{.stl}, \texttt{.obj}), the geometry is imported into FreeCAD, converted to a solid body, and processed using a custom macro that uniformly samples the interior of the object at a chosen spacing $dx$. The material parameters (density, Young’s modulus, Poisson’s ratio) are then assigned to obtain a unified digital model compatible with SPH solvers.

The lower portion of the figure illustrates the standardized grasping protocol used for SPH experiments: particles are initialized with the gripper fingers open, followed by a squeezing phase, a brief hold, and finally a lifting phase used to assess grasp stability and deformation behaviour. Using the same preprocessing and control pipeline ensures that different SPH implementations operate on identical particle sets under comparable manipulation conditions.

The resulting particle clouds were simulated using Smoothed Particle Hydrodynamics (SPH) through two complementary implementations: \textbf{PySPH} \cite{ramachandran2021a, pysph_docs} and \textbf{Taichi} \cite{taichi_overview_docs}. In both cases, deformable bodies are represented as interacting particles that approximate continuum behaviour. Field quantities such as mass density and pressure are reconstructed via kernel interpolation and updated over time using explicit time stepping. 
%This particle-based formulation naturally accommodates large deformations and potential topological changes without the need for re-meshing.

PySPH is an open-source Python framework that provides SPH operators for continuum simulation on a CPU backend \cite{pypr_pysph_github}. In our setup, frictional boundary conditions were defined for gripper--object interactions. However, several operators rely on CPU-based linear algebra routines (e.g., eigen-decomposition), which preclude GPU execution through the OpenCL backend; consequently, all PySPH simulations were executed on the CPU using OpenMP. PySPH does not provide real-time visualization during simulation; instead, individual frames are written to an output directory and later assembled in Mayavi for 3D visualization and per-particle analysis. This post-processing workflow enables detailed diagnostics, but limits interactive performance.

Taichi can JIT-compile Python kernels to GPU code (e.g., CUDA) and provides efficient particle updates with minimal Python overhead. We implemented a weakly-compressible SPH (WCSPH) pipeline as Taichi kernels, including grid-based neighbor search (cell size $h$), density estimation, pressure forces, and artificial viscosity. In the Taichi implementation, the density is calculated by kernel summation using a Poly6 kernel, and pressure forces are calculated using the Spiky kernel gradient. Pressure is defined via a Tait-style equation of state
\begin{equation}
p_i = \frac{c_0^2\rho_0}{\gamma}\left[\left(\frac{\rho_i}{\rho_0}\right)^{\gamma}-1\right],
\end{equation}
with exponent $\gamma$ and reference density $\rho_0$. The sound speed $c_0$ is derived from the bulk modulus $K=E/[3(1-2\nu)]$ via $c_0=\sqrt{K/\rho_0}$ to relate the WCSPH compressibility level to the target material parameters. To further reduce residual compressibility under contact, we apply a lightweight density-projection step that corrects particle positions along local density gradients after force application. 

% This configuration enabled interactive high-throughput execution with near-real-time visualization of particle motion, while periodically snapshotting arrays to the host system for analysis.

In addition to the baseline Taichi SPH formulation, we implemented an ablation that augments particle dynamics with a lightweight mass--spring regulariser. Starting from the same volumetric particle discretisation, we pre-compute an undirected connectivity graph $\mathcal{E}$ in the rest configuration by connecting each particle to up to $k$ nearby particles within a cutoff radius $r_{\mathrm{cut}}$ proportional to $dx$ (in our implementation $k=26$ and $r_{\mathrm{cut}}=1.8\,dx$). 
% For each edge $(i,j)\in\mathcal{E}$ we store a rest length $L_{ij}=\|\mathbf{x}_i^0-\mathbf{x}_j^0\|$ and apply a spring--damper force
% \begin{equation}
% \mathbf{f}_{ij} =
% k_{ij}\left(\|\mathbf{x}_j-\mathbf{x}_i\| - L_{ij}\right)\mathbf{d}_{ij}
% + c_{ij}\left((\mathbf{v}_j-\mathbf{v}_i)\cdot \mathbf{d}_{ij}\right)\mathbf{d}_{ij},
% \end{equation}
% where $\mathbf{d}_{ij}=(\mathbf{x}_j-\mathbf{x}_i)/\|\mathbf{x}_j-\mathbf{x}_i\|$. The net spring contribution to particle $i$ is $\mathbf{f}_i^{\mathrm{MSM}}=\sum_{(i,j)\in\mathcal{E}}\mathbf{f}_{ij}$, and the resulting acceleration $\mathbf{a}^{\mathrm{MSM}}_i=\mathbf{f}_i^{\mathrm{MSM}}/m$ is added to the baseline SPH acceleration prior to integration:
% \begin{equation}
% \mathbf{a}_i = \mathbf{a}^{\mathrm{SPH}}_i + \mathbf{a}^{\mathrm{MSM}}_i + \mathbf{g}.
% \end{equation}
The stiffness of the spring is calibrated from the target elastic parameters by computing the shear modulus $\mu=E/[2(1+\nu)]$ and setting the nominal spring constant proportional to $\mu\,dx$, with a gain factor to tune the effective shear response. 
% Spring damping is chosen as a fixed fraction of critical damping $c_{\mathrm{crit}}=2\sqrt{k m}$ (we use $0.7\,c_{\mathrm{crit}}$). To improve robustness in the contact region, springs whose endpoints lie within the gripper contact band are assigned a modestly higher effective stiffness. 
The MSM term is not intended to replace the SPH constitutive update; it acts as an elastic stabilisation term that increases cohesion and reduces unphysical particle separation during grasping.

% In the Taichi simulations, the gripper is modelled as two parallel plates. Plate motion is time-driven: the plates move inward for a fixed duration (closing), hold briefly (stabilisation), and then lift upward with a ramped vertical velocity. The contact between the particles and the plates is modeled with a penalty-based normal force with damping, combined with Coulomb friction (distinct static and kinetic coefficients). A small Baumgarte-style positional correction is applied to reduce interpenetration. This time-driven protocol yields repeatable kinematics across runs, allowing controlled comparisons between baseline SPH and SPH+MSM under identical grasp trajectories.

To ensure a controlled comparison, SPH and SPH+MSM use identical particle sets, boundary conditions, friction settings, and time-stepping parameters; only the additional spring--damper term (and its hyperparameters) differs between the two Taichi runs. All SPH experiments were performed on a \emph{Legion Pro 7i Gen 10} laptop (Intel Core Ultra 9 275HX CPU; NVIDIA GeForce RTX~5080 GPU with 16\,GB VRAM). 

The SPH and FEM pipelines were configured to achieve broadly comparable conditions rather than exact parity. Geometry, nominal material parameters ($\rho$, $E$, $\nu$), friction coefficient, gripper movement, and lift targets were matched as closely as practicable. Solver-specific parameters were adjusted only as necessary for numerical stability, such as contact tolerances in FEM or smoothing length, artificial viscosity, and density projection strength in SPH. Consequently, the resulting comparisons reflect method-level rather than code-level differences under harmonized yet distinct simulation environments.

All quantitative results, together with their detailed analysis, are provided in the supplementary material. To support reproducibility, all scripts and macros used to generate volumetric point clouds in FreeCAD, together with the PySPH and Taichi simulation code used in this study, will be released in a public GitHub repository upon publication. 
% The repository will include the preprocessing pipeline, grasping setups, and experiment configurations to facilitate reproducibility and further research.

\section{Results and Discussion} \label{results}
\subsection{Comparative results across setups (FEM vs SPH)}

We compared FEM-based simulations in DefGraspSim with SPH-based simulations using PySPH and Taichi in several representative test cases. A series of parameter sweeps was conducted to examine how variations in material properties influenced the behaviour of both FEM and SPH models during grasping. Three key constitutive parameters were evaluated: Young’s modulus ($E$), Poisson’s ratio ($\nu$) and density ($\rho$). Across these tests, the two formulations demonstrated markedly different sensitivity ranges and numerical stability limits.

\begin{figure*}[!t]
    \centering
    \subfloat[Real grasp of tofu \cite{huang2022defgraspsim}]
    {
        \includegraphics[width=0.32\columnwidth]{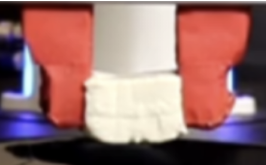}
        \label{fig:tofu}
    }
    \hfil
    \subfloat[FEM simulation (DefGraspSim) \cite{huang2022defgraspsim}]
    {
        \includegraphics[width=0.32\columnwidth]{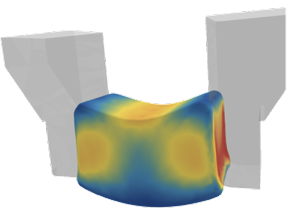}
        \label{fig:DefGraspSim_tofu}
    }
    \hfil
    \subfloat[SPH simulation using Taichi]
    {
        \includegraphics[width=0.32\columnwidth]{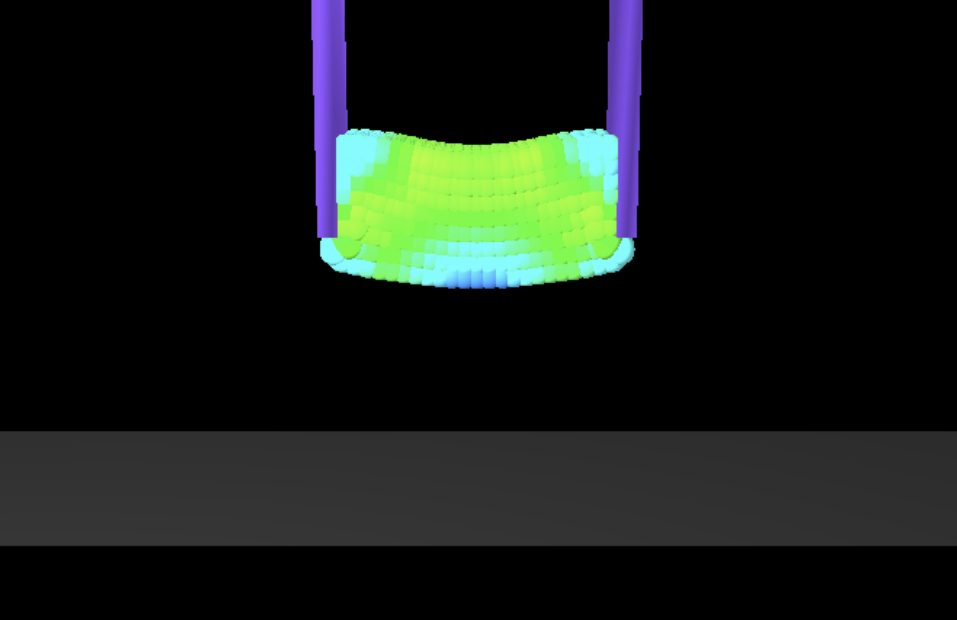}
        \label{fig:taichi_tofu}
    }
    \caption{Comparison of tofu grasping across real-world experiments, FEM, and SPH simulation frameworks.}
    \label{fig:tofu_framework_comparison}
\end{figure*}

\begin{figure*}[t]
    \centering
    % Optional: reduce horizontal spacing between images
    \setlength{\tabcolsep}{2pt}
    \begin{tabular}{cccc}
        % Top row: Taichi with Mass Spring
        \includegraphics[width=0.20\textwidth]{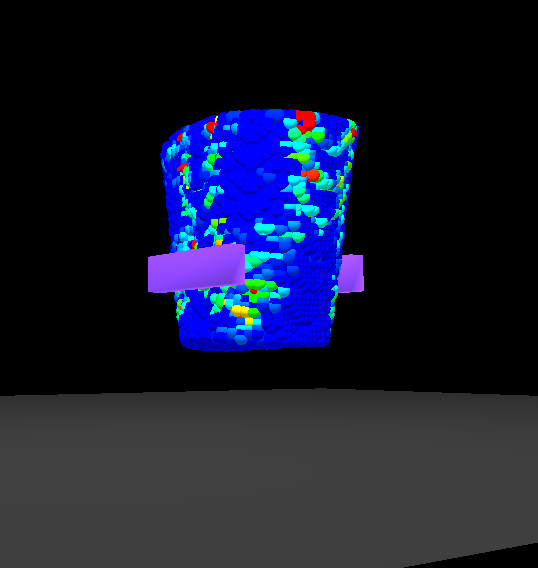} &
        \includegraphics[width=0.20\textwidth]{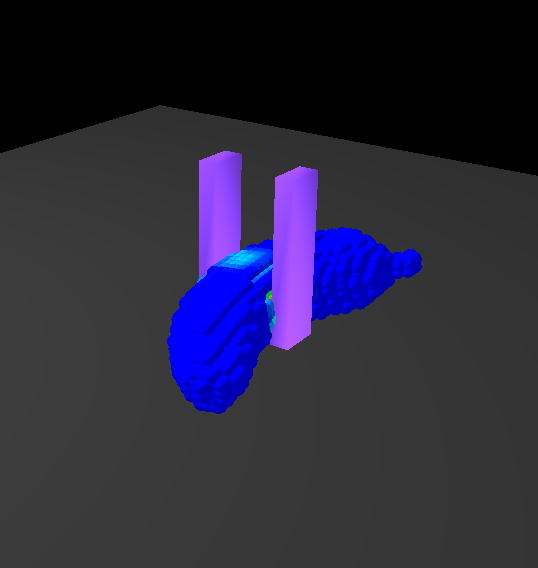} &
        \includegraphics[width=0.20\textwidth]{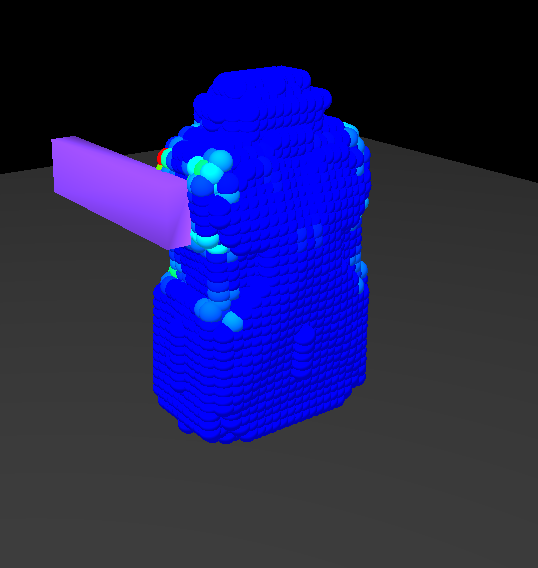} &
        \includegraphics[width=0.20\textwidth]{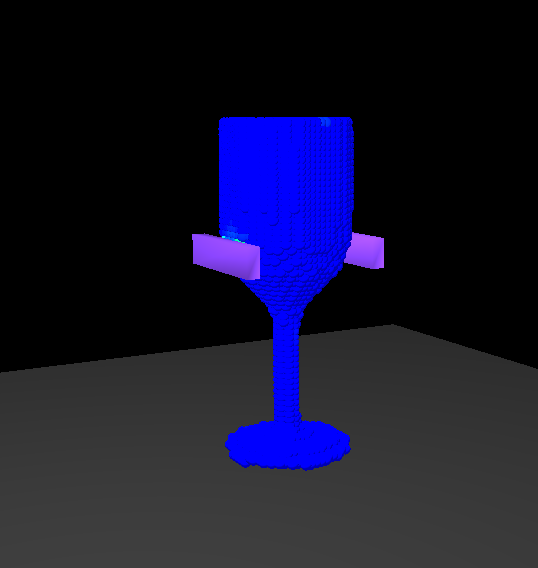} \\
        % Middle row: Taichi no Spring
        \includegraphics[width=0.20\textwidth]{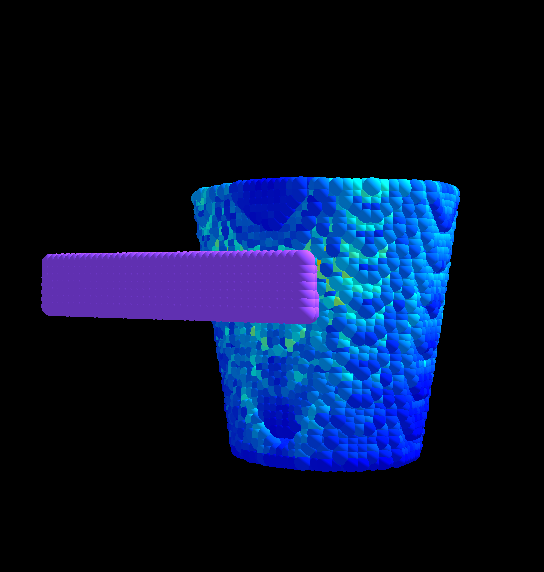} &
        \includegraphics[width=0.20\textwidth]{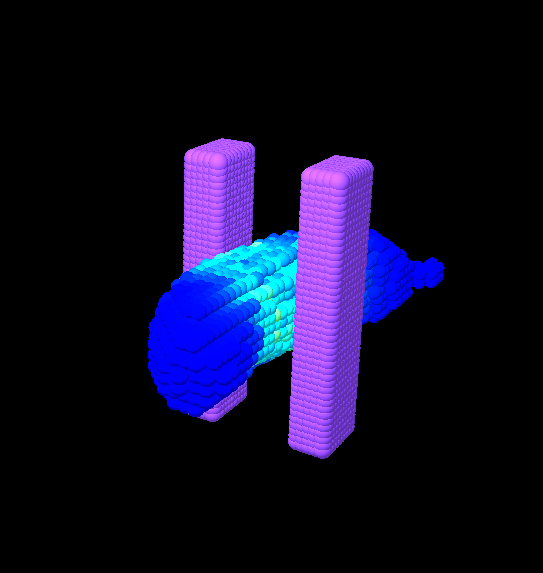} &
        \includegraphics[width=0.20\textwidth]{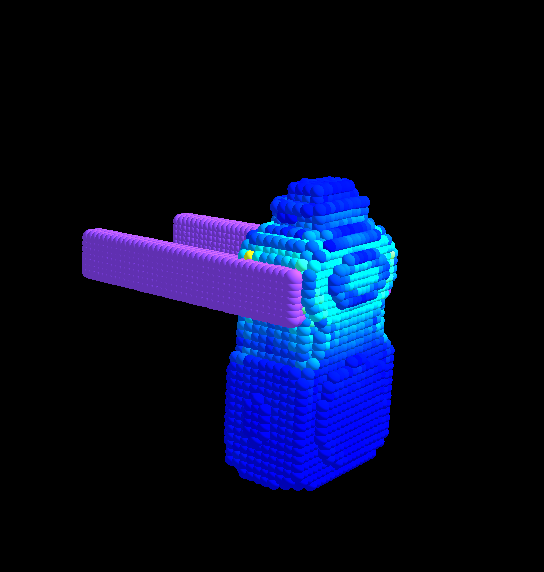} &
        \includegraphics[width=0.20\textwidth]{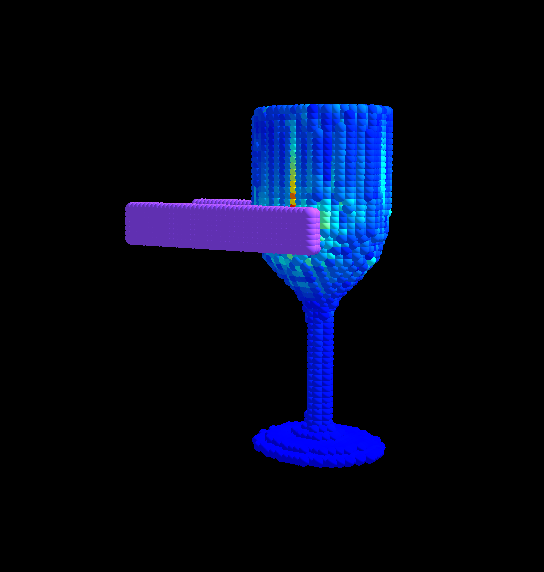} \\
        % Bottom row: PySPH
        \includegraphics[width=0.20\textwidth]{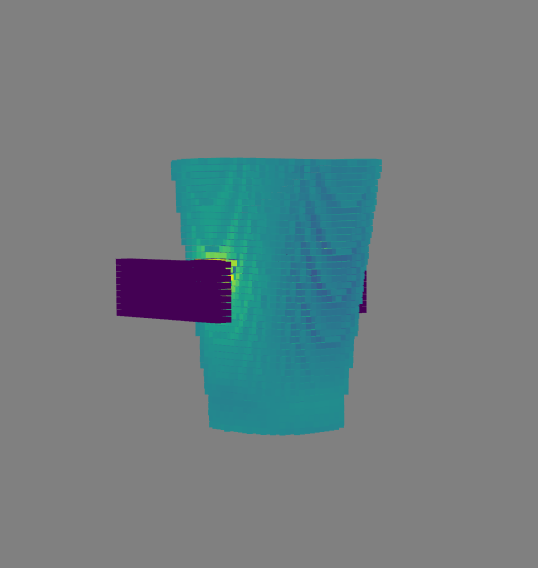} &
        \includegraphics[width=0.20\textwidth]{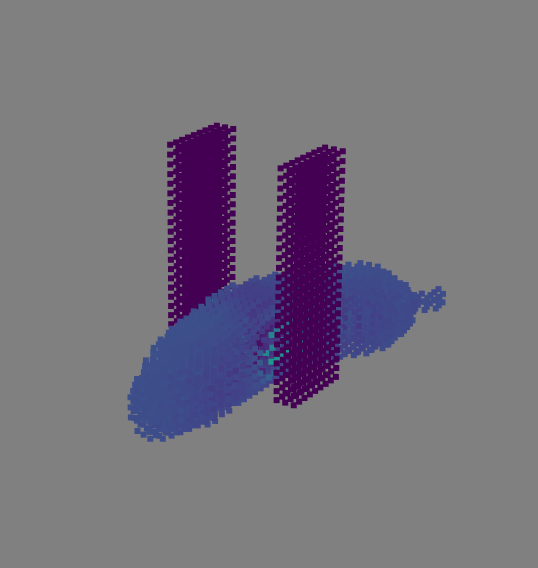} &
        \includegraphics[width=0.20\textwidth]{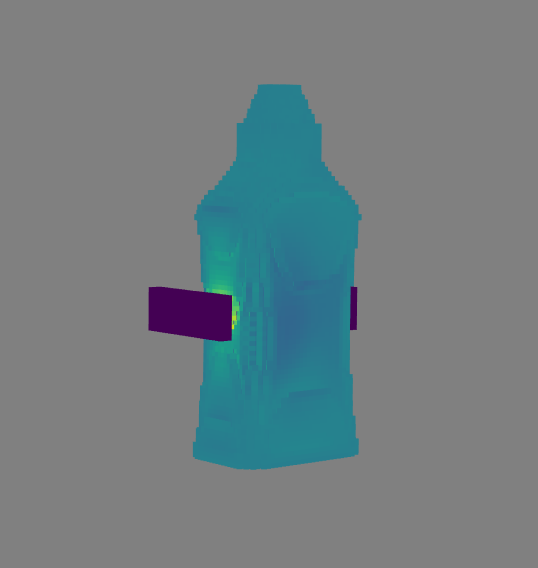} &
        \includegraphics[width=0.20\textwidth]{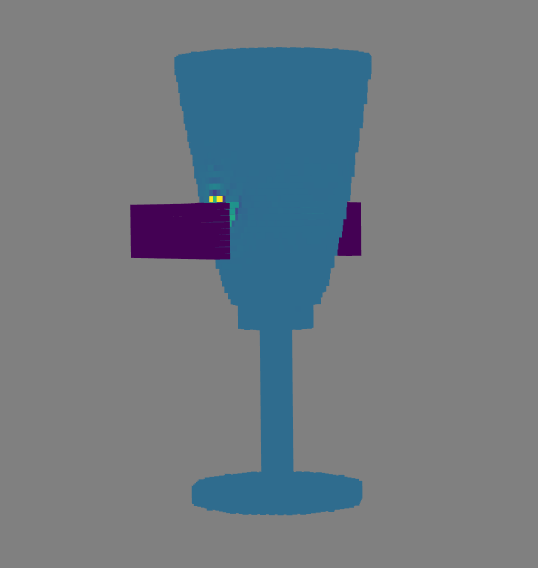} \\
    \end{tabular}
    \caption{Qualitative comparison of grasp simulations for four objects
    (plastic cup, banana, bleach bottle, and wine glass) using Taichi
    (top row) and PySPH (bottom row). From left to right: plastic cup,
    banana, bleach bottle, wine glass.}
    \label{fig:sim_grid}
\end{figure*}

Note that pickup-force values are reported as measured by each simulator under the same high-level squeeze--hold--lift protocol, but are not strictly comparable across methods because the object geometries are not identical. In particular, the exact object dimensions used in DefGraspSim are not fully specified, so we matched shapes by proportions and nominal scaling, but small differences in absolute size remain. As a result, pickup-force magnitudes should be interpreted within each simulator (trends across parameter sweeps) rather than as a direct cross-simulator force benchmark.

\begin{figure*}[h!]
\centering
\begin{tabular}{cccc}
  \includegraphics[width=0.22\linewidth]{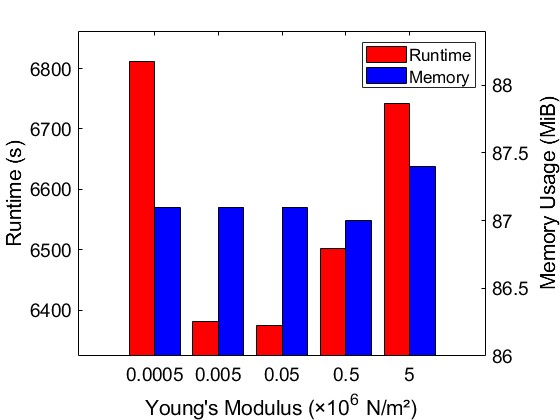} &
  \includegraphics[width=0.22\linewidth]{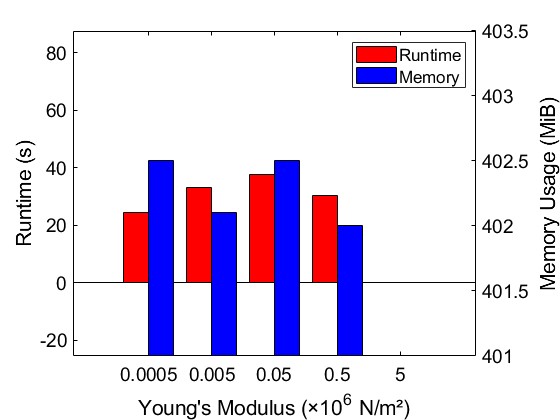} &
  \includegraphics[width=0.22\linewidth]{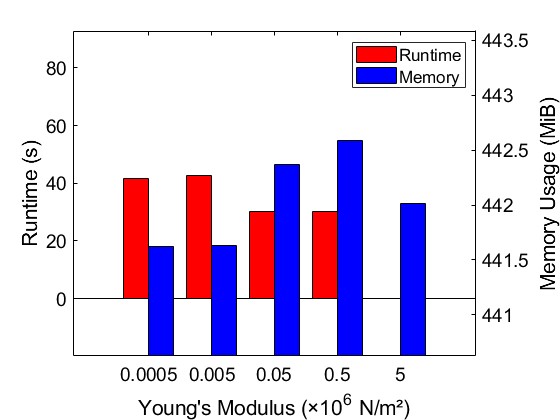} &
  \includegraphics[width=0.22\linewidth]{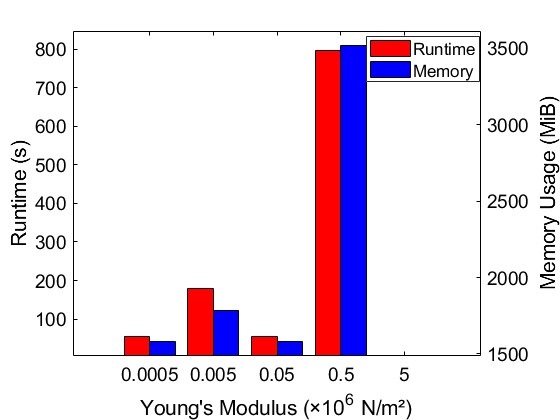} \\
   (i) & (ii) & (iii) & (iv)
\end{tabular}
\caption{Young’s modulus sweep at constants $\rho = 1000~\mathrm{kg/m^3}$, $\nu = 0.3$ (Poisson’s ratio), and $\mu = 0.7$ (friction coefficient), using different simulation methods: (i) PySPH, (ii) Taichi, (iii) Taichi with mass-spring, and (iv) DefGraspSim.}
\label{ym}
\end{figure*}

\textbf{Young’s modulus} was swept from \(E=500\) to \(5\times10^{6}\,\mathrm{Pa}\), while holding \(\rho=1000\,\mathrm{kg\,m^{-3}}\), \(\nu=0.3\), and \(\mu=0.7\) fixed. Simulations with PySPH succeed only at the highest stiffness $E=5\times10^{6}\,\mathrm{Pa}$.
%This pattern is consistent with a cohesion/shape-retention limitation at low stiffness: the body deforms excessively under the closing plates, leading to slip or particle separation before a stable lift is achieved. 
They also have the highest computational cost in this sweep (approximately $6.4$ -- $6.8\times10^{3}$ \,s per run, i.e., $\sim$1.8 \, hours) but relatively low memory growth ($\sim$87 \, MiB), reflecting its CPU-bound execution and diagnostic pipeline (Fig.\ref{ym}).

The Taichi baseline SPH matched PySPH qualitatively, while being markedly more efficient (24--38s per run; $\sim$402 \,MiB memory). Successful pickups were not obtained for any $E$ in this sweep (Fig.\ref{ym}). And, the absence of data point at $E=5\times10^{6}$ was due to stiffness-induced instability, which required a reduced time step to achieve stable simulation and to achieve a successful pick up as with PySPH. Taichi SPH+MSM (spring-regularized SPH, Fig.\ref{ym}) changes the picture by introducing an explicit elastic network that resists separation and improves shape retention under contact. SPH+MSM remains unsuccessful at the very soft end ($E=500$ and $5000\,\mathrm{Pa}$), but succeeds at intermediate stiffness ($E=5\times10^{4}$ and $5\times10^{5}\,\mathrm{Pa}$), where both PySPH and baseline Taichi SPH fail. This supports the intended function of the spring layer as a cohesion/shape-retention regulariser, extending the range of $E$ values for which particle-based grasping is stable under plate contact.

% Taichi baseline SPH is computationally more efficient (24--38 \,s per run; $\sim$402 \,MiB memory) but does not achieve successful pickups at any tested $E$ in this sweep (Table~\ref{tab:taichi_youngs_results}). In practice, this indicates that the baseline particle formulation, under the chosen contact/friction and time-stepping settings, lacks sufficient tensile/cohesive resistance for this grasping protocol: increasing $E$ in the nominal material parameters alone does not reliably translate into grasp-stable behaviour in the GPU SPH pipeline. The absent data point at $E=5\times10^{6}$ is consistent with stiffness-induced instability, which implies that a reduced time step would be needed for stable simulation and to achieve a successful pick up as with PySPH.

%However, the pickup-force proxy increases sharply between $E=5\times10^{4}$ (46.6 \,N) and $E=5\times10^{5}$ (364.5 \,N), indicating that the effective stiffness of the hybrid model can become dominated by the spring network (or its interaction with contact), rather than smoothly scaling with continuum parameters. 
At the highest stiffness tested($E=5\times10^{6}$), SPH+MSM becomes unstable again, consistent with the instability produced by the high stiffness, which can be solved by lowering the simulation time step. FEM (using DefGraspSim) shows a similar trend as SPH+MSM: it succeeds primarily in the mid-range and fails at the extremes. In particular, DefGraspSim succeeds for $E=5\times10^{3}$--$5\times10^{5}\,\mathrm{Pa}$ but fails for $E=500\,\mathrm{Pa}$ and fails at $E=5\times10^{6}\,\mathrm{Pa}$. Both runtime and memory grow substantially with stiffness (e.g., 54 \,s and $\sim$1.6 \,GiB at $5\times10^{4}$ vs 797 \,s and $\sim$3.5 \,GiB at $5\times10^{5}$, Fig.\ref{ym}), consistent with increased solver difficulty in rigid, contac-rich regimes. The reported pickup forces also show a strong outlier at $E=5000\,\mathrm{Pa}$ (11.4 \,kN), while forces at $E\ge 5\times10^{4}$ are in a plausible range (6--7 \,N); this discrepancy was caused by the sliding edge pickup behaviour. %This suggests that the force metric is sensitive to contact/penetration transients and/or control termination conditions at low stiffness, and should be interpreted together with failure mode labels and force-time traces rather than as a single scalar alone.

% Overall, the sweep indicates that (i) FEM can be accurate and stable in a mid-stiffness regime but becomes difficult to converge at very high stiffness; (ii) baseline GPU SPH is fast but, without additional stabilization, is prone to cohesion/retention failures under grasping contact; and (iii) SPH+MSM bridges this gap by enabling successful grasps at intermediate stiffness, at the cost of introducing an additional stiffness pathway that can dominate the effective response and requires careful calibration to maintain physical interpretability.

\begin{figure*}[h!]
\centering
\begin{tabular}{cccc}
  \includegraphics[width=0.22\linewidth]{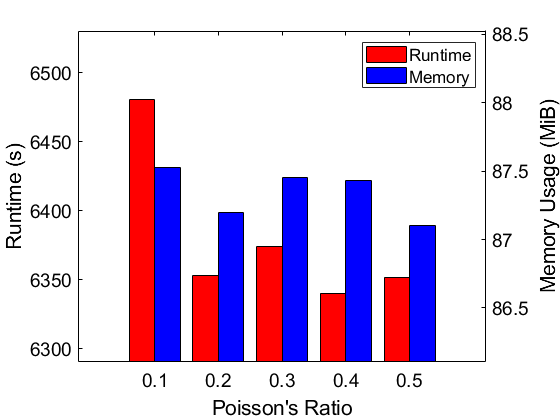} &
  \includegraphics[width=0.22\linewidth]{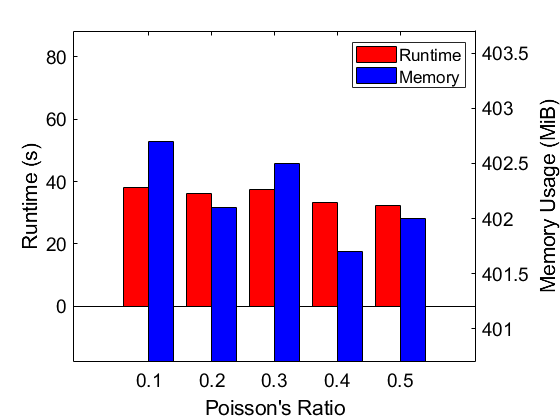} &
  \includegraphics[width=0.22\linewidth]{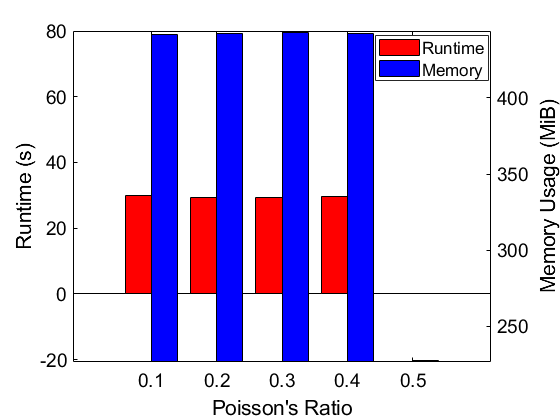} &
  \includegraphics[width=0.22\linewidth]{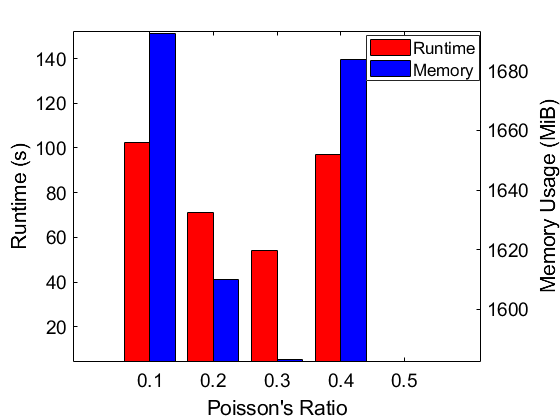} \\
   (i) & (ii) & (iii) & (iv)
\end{tabular}

\caption{Simulation results for various Poisson’s ratios using different methods: (i) PySPH, (ii) Taichi, (iii) Taichi with mass-spring, and (iv) DefGraspSim.}
\label{pr}
\end{figure*}

\textbf{Poisson’s ratio} was varied from $\nu=0.1$ to $0.5$ at fixed density $\rho=1000~\mathrm{kg\,m^{-3}}$, Young’s modulus $E=5\times10^{4}~\mathrm{Pa}$, and friction coefficient $\mu=0.7$. Both PySPH and Taichi baseline SPH achieved zero successful pickups for all tested Poisson ratios. The Taichi baseline remains computationally efficient (32--38 \,s per run, $\sim$402 \,MiB, (Fig.\ref{pr})), while PySPH is orders of magnitude slower ($\sim$6.3--6.5$\times10^{3}$\,s per run, $\sim$87 \,MiB, Fig.\ref{pr}), but neither formulation achieves grasp stability. These results suggest that, in this very soft $E$ regime, varying $\nu$ alone is insufficient to recover grasp stability in the baseline particle formulation; the limiting factor is inadequate cohesion/shape retention under contact rather than stiffness-driven numerical instability. 

% This setting corresponds to a highly compliant, gel-like material, so grasp success depends strongly on whether the simulator maintains sufficient cohesion/shape retention under contact during the squeeze--hold--lift protocol.

%\textbf{SPH+MSM succeeds for $\nu=0.1$--$0.4$ but fails at $\nu=0.5$.}
Adding the spring regulariser markedly changes the outcome. SPH+MSM achieves successful pickup for $\nu=0.1$--$0.4$ (4/5 successes), with runtime comparable to baseline Taichi ($\sim$29--30 \,s, Fig.\ref{pr}) and only a modest increase in memory usage ($\sim$442 \,MiB, Fig.\ref{pr}). The pickup force decreases slightly as $\nu$ increases from 0.1 to 0.4 (50.2\,N $\rightarrow$ 45.7\,N), consistent with a small reduction in effective resistance under the combined particle+spring model. In the nearly incompressible setting $\nu=0.5$, the run terminates due to a numerical singularity in the current parameterization; using $\nu=0.499$ avoids this issue and yields a successful pickup.

%FEM is successful for $\nu=0.1$--$0.4$ but times out at $\nu=0.5$.
DefGraspSim shows consistent success for $\nu=0.1$ -- $0.4$ with pickup forces in a narrow and plausible range (6.97--7.86 \,N) and with a run time varying between 54 \,s and 102 \,s (Fig.\ref{pr}). At $\nu=0.5$, DefGraspSim stalls before achieving a successful pick up, consistent with the known challenges of standard displacement-based FEM when approaching incompressibility, where locking and ill-conditioning can degrade convergence unless mixed formulations, selective integration, or near-incompressible stabilization are used.

% In this soft-material regime ($E=5\times10^{4}\,\mathrm{Pa}$), the Poisson sweep demonstrates that (i) baseline SPH (both CPU and GPU implementations) does not achieve grasp stability for any $\nu$ tested, implying that cohesion/shape retention under contact is the limiting factor; (ii) adding a sparse spring network enables successful pickup over a broad range of $\nu$ (0.1--0.4), indicating that the mass--spring layer effectively supplies missing tensile/cohesive structure; and (iii) both FEM and SPH+MSM exhibit a failure mode at $\nu=0.5$, albeit likely for different numerical reasons (FEM near-incompressibility conditioning vs time step stability in SPH+MSM).

%This sweep probes inertial effects (particle/element mass) and how each solver’s contact handling and damping respond to increased mass under the same squeeze--hold--lift trajectory.

\textbf{Density} was swept from $\rho=800$ to $1200~\mathrm{kg\,m^{-3}}$ with fixed $E=5\times10^{4}~\mathrm{Pa}$, $\nu=0.3$ and $\mu=0.7$. As the Poisson sweep, both PySPH and Taichi baseline SPH formulation achieved zero successful pickups throughout the full density range. Taichi baseline remains computationally efficient (32.7--37.5 \,s, $\sim$402 \,MiB), while PySPH is substantially slower ($\sim$6.37$\times10^{3}$-6.42$\times10^{3}$\,s, $\sim$87 \,MiB), but neither baseline configuration attains grasp stability. %The reported pickup-force values in the Taichi baseline remain extremely small and increase only marginally with density (from $6.36\times10^{-2}$\,N at $\rho=800$ to $8.29\times10^{-2}$\,N at $\rho=1200$), indicating that simply increasing mass does not resolve the underlying failure mode (excessive deformation/slip or loss of contact) in the baseline formulation at this low stiffness.

%SPH+MSM is successful across all densities, with pickup force increasing with $\rho$.
In contrast, the spring-regularized variant is successful for all densities tested (5/5 success) with runtimes comparable to baseline Taichi ($\sim$29--35\,s) and a modest memory increase to $\sim$442\,MiB. Unlike baseline SPH, SPH+MSM produces substantial and density-dependent pickup forces. The overall increase with density is consistent with larger inertial resistance and contact reactions under the same imposed squeeze/lift kinematics, while the non-monotonic dip at $\rho=900$ suggests sensitivity to contact transients, spring-network tension distribution, or the specific force-aggregation definition used for this metric. Crucially, however, the grasp success remains robust: the added elastic network provides sufficient cohesion/shape retention so that density changes do not induce failure.

%FEM remains robust and produces consistent forces across $\rho$.
DefGraspSim also succeeds for all densities (5/5) and yields pickup forces within a narrow band (7.10--8.33 \, N), indicating limited sensitivity of grasp stability and peak force to density in this quasi-static manipulation regime. Runtime varies substantially (58--143 \, s) and memory usage is considerably higher than the Taichi variants (1.59--1.80 \,GiB), reflecting the cost of solving the implicit dynamics of FEM with contact.

% For this soft-material setting ($E=5\times10^{4}\,\mathrm{Pa}$), density variation alone does not recover grasp success in baseline SPH, whereas adding a sparse spring network makes the particle method robust across $\rho=800$--$1200~\mathrm{kg\,m^{-3}}$. FEM is also robust across the density range, producing comparatively stable pickup-force values, but with higher compute and memory demands. In general, the density sweep reinforces the earlier conclusion that the dominant failure mode for baseline SPH in these grasps is insufficient retention of cohesion/shape under contact rather than an adverse choice of $\rho$.

\subsection{Runtime and Scaling of SPH}
All SPH experiments with PySPH were executed on the CPU. The solid–mechanics operators used in our configuration invoke helper routines (such as eigen–decomposition) that are not implemented in PySPH’s OpenCL backend; therefore, the CPU backend was employed for all runs. This setup prioritizes portability and reproducibility over peak performance. %Under fixed hardware, thread count, and random seeds, CPU simulations yielded consistent and repeatable results across repetitions.

To evaluate how computational cost scales with resolution and to identify the discretisation required for reliable grasping, the swept object was represented with particle counts ranging from 351 to 9{,}927. The full quantitative results are reported in the supplementary material. A practical resolution threshold for consistent grasping performance was observed at roughly ${\sim}900$ particles. Beyond this range, the total runtime increased rapidly, reflecting the combined overhead of neighbor-search and increased memory traffic at higher particle counts. The largest simulation, with 9{,}927 particles, required approximately $1.3\times10^{4}\,\mathrm{s}$ (${\approx}3.6$\,h) on a single CPU core.

We also evaluated OpenMP scaling at two representative resolutions: a smaller configuration with ${\sim}1.2\times10^{3}$ particles and a larger configuration with ${\sim}1.0\times10^{4}$ particles. For the smaller configuration, limited computational workload per thread meant that synchronization and cache overheads dominated, yielding only modest speed-ups up to eight threads before performance began to decline at higher counts. In contrast, the larger configuration benefits more clearly from parallel execution, reducing wall-clock time from ${\sim}13{,}000$\, on a single core to ${\sim}4{,}600$–${\sim}4{,}900$\, across 8–16 threads—a speed-up ${\sim}2.7$ –${\sim}2.8\times$ with efficiency ${\sim}18$–${\sim}33\%$. Beyond this range, gains diminish as memory bandwidth and neighbor-search costs become the primary bottlenecks, consistent with the strong-scaling limits typical of particle-based methods.

Although the above results reflect the CPU-bound performance of PySPH, the Taichi-based SPH implementation was substantially faster on the same machine. For a banana grasp scenario with ${\sim}3{,}000$ particles, a complete Taichi run (squeeze + lift) was completed in ${\sim}100$~s on an RTX~5080 GPU, compared to ${\sim}3400$~s for PySPH on a single CPU core. This speed-up enables near real-time visual feedback during interactive experimentation. 

In practice, most Taichi simulations were completed in 0.5 to 5 minutes, depending on the particle count, the chosen time step, and the number of neighbor searches per step. As expected, resolution, stability, and runtime are tightly coupled: increasing particle count improves the fidelity of the object model, but typically requires smaller time steps to maintain numerical stability and avoid particle blow-up, which in turn increases the overall wall-clock time.

\section{Conclusion and Future Work}

This paper compared FEM (using the DefGraspSim framework) with SPH in two implementations (PySPH on the CPU and a GPU Taichi pipeline), and evaluated a spring-regularized variant (SPH+MSM) that overlays a sparse mass--spring network on the SPH particle set to improve cohesion under contact. Across representative objects (tomato, apple, tofu, gelatin, and tube) and systematic sweeps over $(E,\nu,\rho)$, the methods exhibited distinct stability envelopes and failure modes that matter directly for robotic grasping.

Overall, FEM based simulations using DefGraspSim was the most reliable option for mid-range stiffness and moderate compressibility: it produced consistent pickups and plausible forces for structured, mesh-friendly cases (e.g., apple and tube), but its fixed connectivity prevents fracture/tearing and can yield unrealistic ``success'' modes in fragile cases (e.g., corner-catch lifting with extreme forces). FEM also became difficult to solve at the extremes, timing out at very high stiffness and near-incompressibility ($\nu\rightarrow 0.5$). Baseline SPH captured larger deformation and slip-like behaviour more naturally, but in our pinch-grasp protocol, it frequently lost shape retention and failed to maintain load-bearing contact in the low-to-moderate stiffness regimes tested. Adding the MSM regulariser substantially changed this outcome: SPH+MSM enabled successful grasps over a much wider range (notably intermediate $E$ and $\nu\le 0.4$), while preserving the mesh-free contact advantages of SPH. The cost is reduced parameter transparency: the effective response is influenced by spring connectivity, stiffness, and damping, so calibration is required if quantitative force accuracy is the goal rather than qualitative grasp stability. From a computational perspective, the GPU-based Taichi implementations of both SPH and SPH+MSM also achieved a noticeable speedup compared to the FEM pipeline in DefGraspSim, enabling most parameter-sweep simulations to be completed in tens of seconds to a couple of minutes rather than the longer runtimes observed in many FEM cases. 

Although every effort was made to align material properties, geometries, and boundary conditions across tool chains, the FEM and SPH implementations used different solvers and numerical schemes. Consequently, the reported comparisons should be interpreted as method-level evaluations rather than implementation-level evaluations. Absolute runtimes and convergence behaviour depend on solver settings and hardware configuration, but the general trends identified here remain valid across platforms.

Future work could focus on: (i) experimental calibration of both continuum parameters and MSM hyper-parameters using indentation or force--displacement measurements, enabling fairer force comparisons and improved physical grounding of SPH+MSM; (ii) improved numerical robustness for stiff/near-incompressible settings, e.g., adaptive time stepping or implicit/constraint-based integration for the spring term, and near-incompressible stabilization strategies for FEM and SPH; and (iii) hybrid and failure-aware modelling, including mesh-free fracture models or FEM--SPH coupling to localize SPH to high-deformation regions while retaining FEM efficiency elsewhere. %A broader benchmark spanning anisotropic/viscoelastic materials and closed-loop grasp control would further clarify the sim-to-real relevance of each modelling choice.

%\bibliographystyle{cas-model2-names}

%\bibliography{cas-refs}
\bibliographystyle{IEEEtran}
\bibliography{ref}

\end{document}

\title{Supplementary Material for\\
\textit{Bridging Stability and Flexibility in Deformable Object Robotics Grasping across FEM, SPH and Spring-Regularized SPH Models}}
\author{Leyi Ye, Stanley C. Obilikpa and John Oyekan}
\maketitle

\setcounter{table}{0}
\renewcommand{\thetable}{S\arabic{table}}

\section*{S1. Overview}
This supplementary material provides additional quantitative and qualitative results supporting the comparisons presented in the main paper. It includes full parameter-sweep tables for Young’s modulus, Poisson’s ratio, and density; additional object-level qualitative comparisons; runtime and scaling measurements; extended discussion of solver-dependent behaviour; and reproducibility notes regarding code and preprocessing tools.

\section*{S2. Additional Quantitative Results}
This section reports the complete numerical results for the parameter sweeps discussed in the main paper. Unless otherwise stated, the same squeeze-hold--lift grasp protocol was used across all simulations.

\subsection*{S2. Young´s Modulus Sweep Results}
Tables~\ref{tab:pysph_youngs_results}--\ref{tab:defgraspsim_youngs_results} report the full results
for the Young's modulus sweep (constants: $\rho=1000$, $\nu=0.3$, $\mu=0.7$).

\begin{table}[ht]
\centering
\small
\begin{tabular}{rrrrrrrrl}
\toprule
Density & Young's & Poisson's & Friction & Pick up & Simulation  & Memory usage  & Pickup  & Failure\\
& & & & force (N) & time (s) & increment during  & success & reason\\
& & & & & & runtime (MiB) & &\\
\midrule
1000 & 500     & 0.3 & 0.7 & -- & 6812.1 & 87.1 & FALSE & object too soft\\
1000 & 5000    & 0.3 & 0.7 & -- & 6380.8 & 87.1 & FALSE & object too soft\\
1000 & 50000   & 0.3 & 0.7 & -- & 6374.5 & 87.1 & FALSE & object too soft\\
1000 & 500000  & 0.3 & 0.7 & -- & 6502.5 & 87.0 & FALSE & object too soft\\
1000 & 5000000 & 0.3 & 0.7 & -- & 6742.6 & 87.4 & TRUE  & -- \\
\bottomrule
\end{tabular}
\caption{Simulation results across different Young's modulus with PySPH}
\label{tab:pysph_youngs_results}
\end{table}

\begin{table}[ht]
\centering
\small
\begin{tabular}{rrrrrrrrl}
\toprule
Density & Young's & Poisson's & Friction & Pick up & Simulation  & Memory usage  & Pickup  & Failure\\
& & & & force (N) & time (s) & increment during  & success & reason\\
& & & & & & runtime (MiB) & &\\
\midrule
1000 & 500     & 0.3 & 0.7 & -- & 24.6 & 402.5 & FALSE & object too soft\\
1000 & 5000    & 0.3 & 0.7 & -- & 33.1 & 402.1 & FALSE & object too soft\\
1000 & 50000   & 0.3 & 0.7 & -- & 37.5 & 402.5 & FALSE & object too soft\\
1000 & 500000  & 0.3 & 0.7 & -- & 30.4 & 402.0 & FALSE & object too soft\\
1000 & 5000000 & 0.3 & 0.7 & --       & --   & --    & FALSE & unstable time-step for high stiffness\\
\bottomrule
\end{tabular}
\caption{Simulation results across different Young's modulus with Taichi}
\label{tab:taichi_youngs_results}
\end{table}

\begin{table}[ht]
\centering
\small
\begin{tabular}{rrrrrrrrl}
\toprule
Density & Young's & Poisson's & Friction & Pick up & Simulation  & Memory usage  & Pickup  & Failure\\
& & & & force (N) & time (s) & increment during  & success & reason\\
& & & & & & runtime (MiB) & &\\
\midrule
1000 & 500     & 0.3 & 0.7 & -       & 41.78 & 441.625       & FALSE & object too soft\\
1000 & 5000    & 0.3 & 0.7 & -       & 42.76 & 441.6328125  & FALSE & object too soft\\
1000 & 50000   & 0.3 & 0.7 & 46.594  & 30.25 & 442.3671875  & TRUE  & -- \\
1000 & 500000  & 0.3 & 0.7 & 364.511 & 30.14 & 442.5859375  & TRUE  & -- \\
1000 & 5000000 & 0.3 & 0.7 & -       & -     & 442.015625   & FALSE & unstable time-step for high stiffness\\
\bottomrule
\end{tabular}
\caption{Simulation results across different Young's modulus with Taichi (with mass-spring).}
\label{tab:taichi_spring_youngs_results}
\end{table}

\begin{table}[!t]
\centering
\small
\begin{tabular}{rrrrrrrrl}
\toprule
Density & Young's & Poisson's & Friction & Pick up & Simulation  & Memory usage  & Pickup  & Failure\\
& & & & force (N) & time (s) & increment during  & success & reason\\
& & & & & & runtime (MiB) & &\\
\midrule
1000 & 500     & 0.3 & 0.7 & -        & 55.190  & 1583.609 & FALSE & object too soft\\
1000 & 5000    & 0.3 & 0.7 & 11375.149& 179.932 & 1787.285 & TRUE  & corner catch\\
1000 & 50000   & 0.3 & 0.7 & 6.974    & 54.237  & 1583.414 & TRUE  & -- \\
1000 & 500000  & 0.3 & 0.7 & 6.585    & 797.076 & 3516.941 & TRUE  & -- \\
1000 & 5000000 & 0.3 & 0.7 & -        & time out & -        & FALSE & not converging\\
\bottomrule
\end{tabular}
\caption{Simulation results across different Young's modulus with DefGraspSim.}
\label{tab:defgraspsim_youngs_results}
\end{table}

\section*{S2.2 Poisson´s ratio sweep results}

Tables~\ref{tab:pysph_poisson_results}--\ref{tab:defgraspsim_poisson_results} report the full results
for the Poisson´s ratio sweep.

\begin{table}[ht]
\centering
\small
\begin{tabular}{rrrrrrrrl}
\toprule
Density & Young's & Poisson's & Friction & Pick up & Simulation  & Memory usage  & Pickup  & Failure\\
& & & & force (N) & time (s) & increment during  & success & reason\\
& & & & & & runtime (MiB) & &\\
\midrule
1000 & 50000 & 0.1 & 0.7 & -- & 6480.3 & 87.5234375  & FALSE & object too soft\\
1000 & 50000 & 0.2 & 0.7 & -- & 6352.9 & 87.1953125  & FALSE & object too soft\\
1000 & 50000 & 0.3 & 0.7 & -- & 6374.1 & 87.453125   & FALSE & object too soft\\
1000 & 50000 & 0.4 & 0.7 & -- & 6340.4 & 87.4296875  & FALSE & object too soft\\
1000 & 50000 & 0.5 & 0.7 & -- & 6351.8 & 87.10546875 & FALSE & object too soft\\
\bottomrule
\end{tabular}
\caption{Simulation results across different Poisson's ratios with PySPH}
\label{tab:pysph_poisson_results}
\end{table}

\begin{table}[!t]
\centering
\small
\begin{tabular}{rrrrrrrrl}
\toprule
Density & Young's & Poisson's & Friction & Pick up & Simulation  & Memory usage  & Pickup  & Failure\\
& & & & force (N) & time (s) & increment during  & success & reason\\
& & & & & & runtime (MiB) & &\\
\midrule
1000 & 50000 & 0.1 & 0.7 & 7.33E-02 & 38.2 & 402.7 & FALSE & object too soft\\
1000 & 50000 & 0.2 & 0.7 & 7.41E-02 & 36.3 & 402.1 & FALSE & object too soft\\
1000 & 50000 & 0.3 & 0.7 & 7.41E-02 & 37.5 & 402.5 & FALSE & object too soft\\
1000 & 50000 & 0.4 & 0.7 & 7.39E-02 & 33.4 & 401.7 & FALSE & object too soft\\
1000 & 50000 & 0.5 & 0.7 & 7.36E-02 & 32.2 & 402.0 & FALSE & object too soft\\
\bottomrule
\end{tabular}
\caption{Simulation results across different Poisson's ratios with Taichi}
\label{tab:taichi_poisson_results}
\end{table}

\begin{table}[!t]
\centering
\small
\begin{tabular}{rrrrrrrrl}
\toprule
Density & Young's & Poisson's & Friction & Pick up & Simulation  & Memory usage  & Pickup  & Failure\\
& & & & force (N) & time (s) & increment during  & success & reason\\
& & & & & & runtime (MiB) & &\\
\midrule
1000 & 50000 & 0.1 & 0.7 & 50.245 & 30.00 & 441.8984375 & TRUE & -- \\
1000 & 50000 & 0.2 & 0.7 & 48.944 & 29.28 & 442.40625   & TRUE & --\\
1000 & 50000 & 0.3 & 0.7 & 46.611 & 29.41 & 442.8671875 & TRUE & --\\
1000 & 50000 & 0.4 & 0.7 & 45.675 & 29.78 & 441.9921875 & TRUE & --\\
1000 & 50000 & 0.5 & 0.7 & -      & -     & 227.640625  & FALSE & zero division error\\
\bottomrule
\end{tabular}
\caption{Simulation results across different Poisson's ratios with Taichi (with mass-spring).}
\label{tab:taichi_spring_poisson_results}
\end{table}

\begin{table}[!t]
\centering
\small
\begin{tabular}{rrrrrrrrl}
\toprule
Density & Young's & Poisson's & Friction & Pick up & Simulation  & Memory usage  & Pickup  & Failure\\
& & & & force (N) & time (s) & increment during  & success & reason\\
& & & & & & runtime (MiB) & &\\
\midrule
1000 & 50000 & 0.1 & 0.7 & 7.862 & 102.352 & 1692.262 & TRUE & --\\
1000 & 50000 & 0.2 & 0.7 & 7.788 & 71.311  & 1610.008 & TRUE & --\\
1000 & 50000 & 0.3 & 0.7 & 6.974 & 54.237  & 1583.414 & TRUE & --\\
1000 & 50000 & 0.4 & 0.7 & 7.732 & 97.056  & 1683.699 & TRUE & --\\
1000 & 50000 & 0.5 & 0.7 & -     & time out & -        & FALSE & locking effect\\
\bottomrule
\end{tabular}
\caption{Simulation results across different Poisson's ratios with DefGraspSim.}
\label{tab:defgraspsim_poisson_results}
\end{table}

\section*{S2.3 Density sweep results}

Tables~\ref{tab:pysph_density_results}--\ref{tab:DefGraspSim_density_results} report the full results
for the density sweep.

\begin{table}[!t]
\centering
\small
\begin{tabular}{rrrrrrrrl}
\toprule
Density & Young's & Poisson's & Friction & Pick up & Simulation  & Memory usage  & Pickup  & Failure\\
& & & & force (N) & time (s) & increment during  & success & reason\\
& & & & & & runtime (MiB) & &\\
\midrule
800  & 50000 & 0.3 & 0.7 & -- & 6370.4 & 87.0 & FALSE & pickup slip\\
900  & 50000 & 0.3 & 0.7 & -- & 6373.9 & 86.7 & FALSE & pickup slip\\
1000 & 50000 & 0.3 & 0.7 & -- & 6374.5 & 87.1 & FALSE & pickup slip\\
1100 & 50000 & 0.3 & 0.7 & -- & 6419.3 & 87.1 & FALSE & pickup slip\\
1200 & 50000 & 0.3 & 0.7 & -- & 6389.9 & 87.3 & FALSE & pickup slip\\
\bottomrule
\end{tabular}
\caption{Simulation results across different densities with PySPH}
\label{tab:pysph_density_results}
\end{table}

\begin{table}[!t]
\centering
\small
\begin{tabular}{rrrrrrrrl}
\toprule
Density & Young's & Poisson's & Friction & Pick up & Simulation  & Memory usage  & Pickup  & Failure\\
& & & & force (N) & time (s) & increment during  & success & reason\\
& & & & & & runtime (MiB) & &\\
\midrule
800  & 50000 & 0.3 & 0.7 & 6.36E-02 & 33.9 & 402.2 & FALSE & pickup slip\\
900  & 50000 & 0.3 & 0.7 & 6.87E-02 & 35.0 & 402.2 & FALSE & pickup slip\\
1000 & 50000 & 0.3 & 0.7 & 7.41E-02 & 37.5 & 402.5 & FALSE & pickup slip\\
1100 & 50000 & 0.3 & 0.7 & 7.83E-02 & 33.2 & 402.5 & FALSE & pickup slip\\
1200 & 50000 & 0.3 & 0.7 & 8.29E-02 & 32.7 & 401.9 & FALSE & pickup slip\\
\bottomrule
\end{tabular}
\caption{Simulation results across different densities with Taichi}
\label{tab:taichi_density_results}
\end{table}

\begin{table}[!t]
\centering
\small
\begin{tabular}{rrrrrrrrl}
\toprule
Density & Young's & Poisson's & Friction & Pick up & Simulation  & Memory usage  & Pickup  & Failure\\
& & & & force (N) & time (s) & increment during  & success & reason\\
& & & & & & runtime (MiB) & &\\
\midrule
800  & 50000 & 0.3 & 0.7 & 27.293 & 34.75 & 442.50391 & TRUE & --\\
900  & 50000 & 0.3 & 0.7 & 18.164 & 31.68 & 442.19531 & TRUE & --\\
1000 & 50000 & 0.3 & 0.7 & 46.386 & 29.9  & 442.17969 & TRUE & --\\
1100 & 50000 & 0.3 & 0.7 & 62.963 & 29.28 & 442.31250 & TRUE & --\\
1200 & 50000 & 0.3 & 0.7 & 64.804 & 29.51 & 442.03516 & TRUE & --\\
\bottomrule
\end{tabular}
\caption{Simulation results across different densities with Taichi (with mass-spring).}
\label{tab:taichi_spring_density_results}
\end{table}

\begin{table}[!t]
\centering
\small
\begin{tabular}{rrrrrrrrl}
\toprule
Density & Young's & Poisson's & Friction & Pick up & Simulation  & Memory usage  & Pickup  & Failure\\
& & & & force (N) & time (s) & increment during  & success & reason\\
& & & & & & runtime (MiB) & &\\
\midrule
800  & 50000 & 0.3 & 0.7 & 7.921 & 143.233 & 1801.878 & TRUE & --\\
900  & 50000 & 0.3 & 0.7 & 8.096 & 140.314 & 1797.930 & TRUE & --\\
1000 & 50000 & 0.3 & 0.7 & 8.329 & 58.275  & 1591.054 & TRUE & --\\
1100 & 50000 & 0.3 & 0.7 & 7.250 & 97.416  & 1685.750 & TRUE & --\\
1200 & 50000 & 0.3 & 0.7 & 7.097 & 72.149  & 1611.359 & TRUE & --\\
\bottomrule
\end{tabular}
\caption{Simulation results across different densities with DefGraspSim.}
\label{tab:DefGraspSim_density_results}
\end{table}

\section*{S3. Additional Qualitative Comparisons}
This section provides further qualitative observations from the object-level experiments, complementing the quantitative comparisons in the main paper.

\subsection*{S3.1 Object-Level Observations}

\begin{figure}[htbp]
    \centering
    \includegraphics[width=0.5\linewidth]{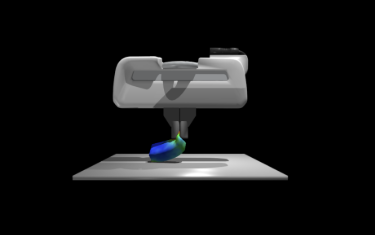}
    \caption{FEM simulation of a rectangular object with tomato material parameters in DefGraspSim, showing the sliding–edge pickup behaviour.}
    \label{fig:DefGraspSim_apple}
\end{figure}

\paragraph{Tomato} The tomato model was expected to rupture once the applied stress exceeded its structural limit. In the FEM simulations, approximately 50\% of the trials achieved pickup, but only after the object first slid downward and was caught in its corner, where the gripper applied unrealistically large forces ($\approx 1.3$--$32~\mathrm{kN}$) to lift it. As the FEM meshes cannot break, the material remained intact even under these extreme loads (see Fig.~\ref{fig:DefGraspSim_apple}). In contrast, all SPH implementations (including MSM+SPH) produced more physically plausible outcomes: successful pickups were obtained only when the object geometry was changed from a rectangle to a sphere, which effectively increased the gripping gap and contact area. 

\paragraph{Apple} The apple model was expected to show primarily elastic deformation, i.e.\ small, reversible shape changes under load. In the FEM simulations, all ten trials were successful, exhibiting negligible visible deformation and grip forces ranging from $11$ to $25~\mathrm{N}$. In SPH, both PySPH and Taichi (with and without springs) achieved successful pickups, with only minor edge damage in the baseline SPH due to localized overstress. 

% However, SPH simulations produced a wider range of outcomes depending on object geometry and gripper configuration. With a spherical geometry and a narrow gripping gap in PySPH, the object initially deformed and then failed before lift, consistent with material rupture once the stress limits were exceeded. When a rectangular geometry and a larger gripping gap were used, both PySPH and Taichi (with and without springs) achieved successful pickups, with only minor edge damage in some cases due to localized overstress.

\paragraph{Tofu} Due to its extreme softness, the gripper must apply a gentle grasp, resulting in large deformations even under small forces. Both the simulated and real experiments with tofu produced highly consistent results (see Fig.~\ref{fig:tofu_framework_comparison}). In all three illustrated cases, real grasp, FEM, and Taichi-based SPH+MSM, tofu was visibly compressed between the fingers, but remained intact and could be lifted, illustrating a good qualitative agreement between numerical and physical behaviour.

\paragraph{Gelatin} For gelatin, FEM and SPH exhibited qualitatively different numerical behaviors but similar physical implications. The FEM simulations did not converge, and excessive stress fields appeared even before gripper contact. In contrast, the SPH (including SPH+MSM) simulations remained stable but showed extensive deformation and no successful lift, consistent with the expected behaviour of very soft, weakly cohesive materials.

\begin{figure*}[!t]
\centering

\subfloat[Real grasp of rubber tube \cite{huang2022defgraspsim}]
{
\includegraphics[width=0.3\linewidth]{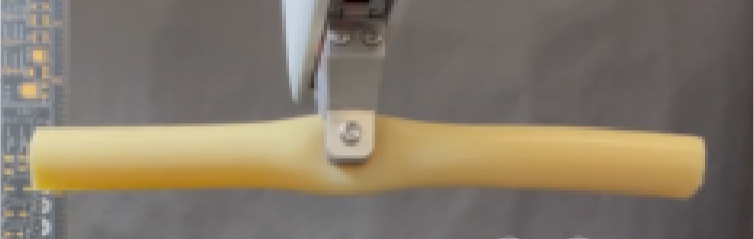}
\label{fig:tube}
}
\hfil
\subfloat[FEM simulation (DefGraspSim) \cite{huang2022defgraspsim}]
{
\includegraphics[width=0.3\linewidth]{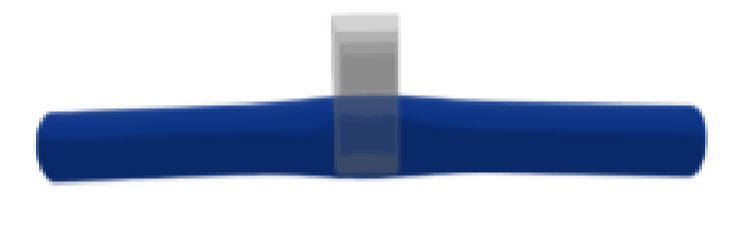}
\label{fig:DefGraspSim_tube}
}
\hfil
\subfloat[SPH simulation using PySPH]
{
\includegraphics[width=0.3\linewidth]{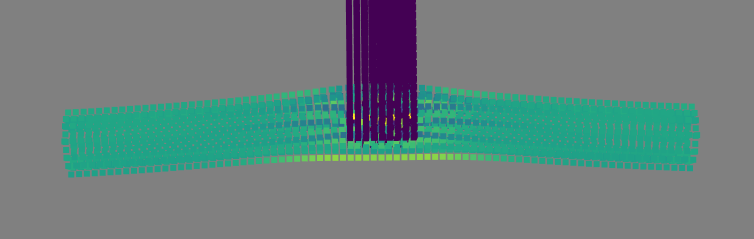}
\label{fig:PySPH_tube}
}

\caption{Qualitative comparison of rubber tube grasping in real-world experiments and in FEM- and SPH-based simulation frameworks.}
\label{fig:tube_framework_comparison}

\end{figure*}

\paragraph{Tube} Real and simulated experiments were also carried out with a rubber tube to evaluate deformation and grasp behaviour. The FEM simulations in DefGraspSim reproduced outcomes consistent with those reported in the reference implementation, showing realistic bending and localized flattening at the contact regions. Both PySPH and Taichi (with and without springs) achieved successful pickups, and the deformation patterns closely matched those observed in the real experiments, demonstrating good qualitative agreement with both FEM and physical experiments, as shown in Fig.~\ref{fig:tube_framework_comparison}.

\subsection*{S3.2 Real--Simulation Comparison}

% \begin{figure*}[!t]
%     \centering
%     \subfloat[Real grasp of rubber tube \cite{huang2022defgraspsim}]
%     {
%         \imgbox[figs/tube2.png]{0.32\columnwidth}
%         \label{fig:tube}
%     }
%     \hfil
%     \subfloat[FEM simulation (DefGraspSim) \cite{huang2022defgraspsim}]
%     {
%         \imgbox[figs/DefGraspSim_tube2.png]{0.32\columnwidth}
%         \label{fig:DefGraspSim_tube}
%     }
%     \hfil
%     \subfloat[SPH simulation using PySPH]
%     {
%         \imgbox[figs/pysph_tube2.png]{0.32\columnwidth}
%         \label{fig:PySPH_tube}
%     }

%     \caption{Qualitative comparison of rubber tube grasping in real-world experiments and in FEM- and SPH-based simulation frameworks.}
%     \label{fig:tube_framework_comparison}
% \end{figure*}

In addition to the tube benchmark, broader qualitative comparisons were also performed across other objects, including a plastic cup, banana, bleach bottle, and wine glass. In these cases, the dominant deformation modes were highly consistent within each category of objects across the different solvers. The color patterns observed in these visualizations should not be interpreted as direct cross-method comparisons of stress magnitude, since the plotted scalar fields are not identical in definition across implementations; depending on the framework, they may correspond to pressure-like proxies, spring strain or tension indicators, or other solver-specific stress surrogates. Nevertheless, visualizations consistently indicate where loads concentrate, typically near finger contact patches, edges, and high-curvature regions, which is useful for diagnosing grasp behaviour and failure modes.
% provides a broader qualitative comparison across additional objects (plastic cup, banana, bleach bottle, and wine glass). The top row corresponds to Taichi SPH+MSM, the middle row corresponds to the Taichi no-spring baseline, and the bottom row corresponds to PySPH. While the \emph{colour patterns} differ across objects and frameworks, this should not be read as a direct cross-method comparison of ``stress magnitude'': the plotted scalar field is not identical in definition across implementations (e.g., pressure-like proxies, spring strain/tension indicators, or solver-specific stress surrogates). Nevertheless, each visualization consistently highlights \emph{where} load concentrates (typically near finger contact patches, edges, and high-curvature regions), which is useful for diagnosing grasp failure. More importantly, across all three methods, the observed deformation modes are highly consistent for each object (compression patterns, bending modes, collapse regions, and overall shape change), indicating agreement on the dominant kinematics of the grasp even when the internal stress proxies differ.

\section*{S4. Runtime and Scaling Details}
This section reports additional computational-performance measurements for the SPH pipelines. To evaluate the relationship between resolution and computational cost, the number of particles in the PySPH object representation was varied from 351 to 9{,}927 particles. As shown in Table~\ref{tbl:particlesweep}, a practical threshold for consistent pickup success emerged around 900 particles. Below this resolution, the object representation was too coarse to maintain stable grasp behaviour. Above this point, runtime increased sharply, reflecting the combined cost of neighbour-search operations and the increased memory traffic associated with higher particle counts.

\begin{table}[t]
  \centering
  \caption{PySPH particle-count sweep on a single-core CPU. Reported are simulation runtime and grasp success. A checkmark indicates successful pickup; a cross indicates failure.}
  \label{tbl:particlesweep}
  \begin{tabular*}{\columnwidth}{@{\extracolsep{\fill}} r r c @{}}
    \toprule
    \#Particles & Simulation time (s) & Pickup \\
    \midrule
    351    & 270.8     & $\times$ \\
    586    & 517.8     & $\times$ \\
    684    & 625.6     & $\times$ \\
    900    & 776.7     & $\checkmark$ \\
    1{,}145 & 1{,}187.5 & $\checkmark$ \\
    6{,}975 & 7{,}689.7 & $\checkmark$ \\
    9{,}927 & 12{,}993.6 & $\checkmark$ \\
    \bottomrule
  \end{tabular*}
\end{table}

\begin{table}[htbp]
  \centering
  \caption{PySPH OpenMP scaling: wall–clock time vs.\ thread count.}
  \label{tbl:openmp}
  \begin{tabular*}{\linewidth}{@{\extracolsep{\fill}} l r r @{}}
    \toprule
    & \multicolumn{1}{c}{\textbf{$\sim$1.2k particles}} & \multicolumn{1}{c}{\textbf{$\sim$10k particles}} \\
    \midrule
    1 thread  & 333.8 s  & 12{,}993.6 s \\
    2 threads & 246.8 s  & 5{,}216.9 s \\
    4 threads & 189.7 s  & 4{,}610.7 s \\
    8 threads & 182.8 s  & 4{,}882.9 s \\
    12 threads & 201.6 s & 4{,}644.8 s \\
    16 threads & 320.9 s & 4{,}611.9 s \\
    24 threads & 727.4 s & 4{,}772.9 s \\
    \bottomrule
  \end{tabular*}
\end{table}

Table~\ref{tbl:openmp} reports the OpenMP scaling behaviour of the PySPH implementation. Parallel execution reduced wall-clock time substantially relative to the single-threaded case, particularly for the larger particle count. However, the speed-up was not monotonic. Beyond a moderate number of threads, overheads associated with synchronisation, scheduling, and memory contention limited further gains and, in some cases, degraded performance. These results suggest that PySPH benefits from moderate CPU parallelisation, but does not scale efficiently to very high thread counts on the tested hardware.

\section*{S5. Extended Discussion}
\subsection*{S5.1 Interpretation of Solver-Dependent Behaviour}

Although we varied material parameters and object types, the squeeze--hold--lift protocol imposes a common requirement across all experiments: the simulation body must maintain load-bearing contact during the hold and lift phases while preserving enough shape retention/cohesion to keep contact normals engaged. Interpreting the results through this lens helps explain why some runs ``fail" even when the solver remains numerically stable: the simulation can produce plausible large deformation, yet still lose the mechanical structure needed for a stable pinch grasp. This also clarifies why pickup success is not, by itself, a sufficient indicator of physical realism. In some regimes, a solver can generate a mechanically stable lift through contact artifacts or constraints that would correspond to rupture/separation in reality, whereas another solver may produce a physically plausible collapse that correctly predicts grasp failure. 

The FEM pipeline generally provides consistent behaviour when the material regime is well-conditioned for implicit contact-rich solves and when the mesh can represent the deformation without requiring connectivity change. However, fixed mesh connectivity constrains the space of representable failure modes. As a result, ``success" can occur in situations where a real object would likely tear, crack, or separate. In those cases, the solver can retain an intact body and still find a force/constraint solution that enables lifting, which should be interpreted as a limitation of the modeling assumptions rather than the evidence of superior grasping performance. Near-incompressible parameterisations introduce an additional difficulty specific to FEM: as $\nu\rightarrow0.5$, the bulk response becomes stiff and standard displacement-based formulations can become ill-conditioned (locking), which manifests as convergence slowdowns or timeouts in contact-rich settings. This explains why the FEM stability envelope tightens near incompressibility even when the high-level grasp protocol remains unchanged. 

Baseline SPH (using PySPH and Taichi) naturally represents large deformation and slip/collapse behaviors without mesh distortion, which is valuable for soft objects under contact. However, pinch grasping is particularly demanding for particle formulations because it requires the body to sustain tensile-like resistance and maintain a coherent structure under concentrated contact patches. In our experiments, varying $\rho$ and $\nu$ does not recover stable pickup in soft regimes, suggesting that the limiting factor is not inertia or compressibility alone, but rather the inability of the baseline configuration to preserve shape retention under sustained pinching contact. In other words, the solver may remain stable while the object mechanically "gives way" in a way that prevents load-bearing contact during lift. 

Overlaying a sparse spring network effectively adds an explicit tensile/shape-retention pathway that counteracts particle separation and excessive collapse. This explains why SPH+MSM expands the region of stable grasps relative to baseline SPH under the same grasp trajectory: the regulariser directly targets the failure mode that dominates soft pinch contact. The cost is reduced parameter transparency: the effective stiffness is now influenced by connectivity, spring calibration, and damping, so the mapping from continuum parameters to observed behaviour becomes less direct unless the spring layer is carefully calibrated.

\subsection*{S5.2 Practical Guidance for Method Selection}

The results suggest that, method choice should be guided less by "which solver is more accurate in general" and more by which behaviors are critical for the task:
\begin{itemize}
    \item FEM would be more appropriate when consistent contact forces and repeatable pickup behaviour are needed in regimes where fracture/topology change is not central, and where near-incompressibility is avoided or handled with special formulations. 
    \item Pure SPH would be more appropriate when large deformation, slip, or separation-like behaviour is expected and when grasp failure itself is a realistic outcome to be predicted (e.g., very soft or weakly cohesive objects).
    \item And SPH+MSM would be appropriate when baseline SPH fails primarily due to cohesion/retention under pinch contact but should be treated as a hybrid model that requires calibration to avoid introducing non-physical tensile stiffness. 
\end{itemize}

Overall, the experiments indicate that FEM is a strong baseline for stable, interpretable grasp simulation in mid-regime materials, while SPH provides qualitatively richer deformation and failure behaviour. Introducing a sparse spring network inside SPH materially expands the grasp-stable parameter range, making SPH more practical for grasp pipelines, but at the cost of additional calibration effort and less direct mapping from $(E,\nu)$ to effective stiffness.

\subsection*{S5.3 Limitations of the Comparison}

The comparison is method-level rather than implementation-level: FEM and SPH differ in time integration, contact handling, and numerical parameters required for stability. In addition, small geometry-scale differences across toolchains affect absolute force magnitudes, so force values are most meaningful within simulator trends rather than direct cross-simulator benchmarks. Stronger cross-method force comparisons would require explicit geometric scale matching and calibration experiments to fit both continuum parameters and SPH+MSM hyper-parameters to the same physical reference. Geometry, nominal material parameters, friction coefficient, gripper motion, and lift targets were matched as closely as practicable. However, small geometry-scale differences across toolchains can still influence absolute force magnitudes. For this reason, force values are most meaningful when interpreted as within-simulator trends rather than as exact cross-simulator benchmarks.

A stronger cross-method force comparison would require explicit geometric scale matching and dedicated calibration experiments to fit both continuum parameters and SPH+MSM hyperparameters to the same physical reference data. Similarly, a more rigorous comparison of failure realism would benefit from experimental benchmarks involving controlled rupture, tearing, or irreversible deformation, which were outside the scope of the present study.

\bibliographystyle{IEEEtran}
\bibliography{ref}